\documentclass[runningheads]{llncs}
\usepackage{enumitem}
\usepackage[T1]{fontenc}
\usepackage{multirow}
\usepackage{amssymb}
\usepackage{amsmath}
\usepackage{graphicx,verbatim}
\usepackage{booktabs}
\usepackage[misc]{ifsym}
\usepackage{pifont}

\usepackage{hyperref}
\hypersetup{colorlinks=true,linkcolor=blue,citecolor=blue,urlcolor=blue}
\begin{document}
\title{Historical Report Guided Bi-modal Concurrent Learning for Pathology Report Generation} 

\author{Ling Zhang\inst{1} \and
Boxiang Yun\inst{1} \and Qingli Li\inst{1} \and Yan Wang\inst{1}\textsuperscript{(\Letter)}}

\institute{{Shanghai Key Laboratory of Multidimensional Information Processing, \\East China Normal University, Shanghai, China}\\
\email{51265904017@stu.ecnu.edu.cn, 52265904012@stu.ecnu.edu.cn, qlli@cs.ecnu.edu.cn, ywang@cee.ecnu.edu.cn}
}
    
\maketitle              
\begin{abstract}
Automated pathology report generation from Whole Slide Images (WSIs) faces two key challenges: (1) lack of semantic content in visual features and (2) inherent information redundancy in WSIs. To address these issues, we propose a novel Historical Report Guided \textbf{Bi}-modal Concurrent Learning Framework for Pathology Report \textbf{Gen}eration (BiGen) emulating pathologists' diagnostic reasoning, consisting of: (1) A knowledge retrieval mechanism to provide rich semantic content, which retrieves WSI-relevant knowledge from pre-built medical knowledge bank by matching high-attention patches and (2) A bi-modal concurrent learning
strategy instantiated via a learnable visual token and a learnable textual token to dynamically extract key visual features and retrieved knowledge, where weight-shared layers enable cross-modal alignment between visual features and knowledge features. Our multi-modal decoder integrates both modals for comprehensive diagnostic reports generation. Experiments on the PathText (BRCA) dataset demonstrate our framework's superiority, achieving state-of-the-art performance with 7.4\% relative improvement in NLP metrics and 19.1\% enhancement in classification metrics for Her-2 prediction versus existing methods. Ablation studies validate the necessity of our proposed modules, highlighting our method's ability to provide WSI-relevant rich semantic content and suppress information redundancy in WSIs. Code is publicly available at https://github.com/DeepMed-Lab-ECNU/BiGen. 


\end{abstract}
\section{Introduction}
The automated analysis of Whole Slide Images (WSIs) in digital pathology has emerged as a significant research direction in the field of medical artificial intelligence recently \cite{el2025whole,lu2021data,xu2024whole}. Among these, the task of pathology report generation based on WSIs has garnered considerable attention due to its immense potential in assisting diagnosis and reducing the workload of physicians \cite{chen2024wsicaption,guo2024histgen,tan2024clinical,tran2024generating}. 

In recent years, there have been a significant number of studies on image captioning \cite{lu2017knowing,vinyals2015show,pmlr-v37-xuc15,zhu2024beyond}. In the medical field, numerous studies have focused on generating radiology reports from radiological images \cite{chen2022cross,chen2020generating,gu2024complex}. There are also studies \cite{li2024llava,lu2024multimodal,tu2024towards} that have focused on obtaining descriptions of small pathological regions of interest. Compared with these report generation tasks, generating reports from WSIs is more challenging and less studied. The challenge relies mainly on how to effectively extract \textbf{semantically diagnostic features} from WSIs with \textbf{ultra-high resolution} \cite{li2023task,zhang2022dtfd}. Existing approaches such as MI-Gen \cite{chen2024wsicaption} directly apply a Transformer model to input image tokens, allowing each token to interact with other tokens. HistGen \cite{guo2024histgen} employs local-global hierarchical visual encoding to input image tokens, which still relies on a self-attention mechanism to explore relationships among multiple tokens. Though global visual features can be learned via self-attention, these methods only learn visual features from WSI itself, which inevitably (1) \textbf{lack of semantic content} and (2) \textbf{introduce considerable redundancy}.

To address these challenges, we propose to leverage historical diagnostic reports to guide the learning of semantically key pathological features for generating current WSI's report. Concretely, we let the visual features be learned along with the relevant textual features, whose knowledge is retrieved from a pre-built knowledge bank. Our method contains two key components: (1) We propose to retrieve WSI-relevant report candidates from a pre-built knowledge bank to provide \textbf{semantically rich} diagnostic report information. (2) We propose a bi-modal concurrent learning strategy. It is instantiated via a learnable visual token and a learnable textual token which extract \textbf{key pathological features} and key knowledge features, respectively. The strategy also enables the learning of visual token to be guided by the semantically rich textual token. The knowledge retrieval emulates the process that pathologists recall historical diagnosis records from similar cases to aid current decision-making. The visual and textual token represent a more global perspective to mirror the way pathologists view WSIs and recall historical diagnosis records—they don't necessarily examine each patch or each record but rather focus on key information to make a comprehensive diagnosis. The main contributions of this paper can be summarized as follows: \begin{itemize}[label=\textbullet, leftmargin=0.5cm]
\item To the best of our knowledge, we first introduce an explicit knowledge retrieval mechanism into the pathology report generation task to enhance semantically rich diagnostic report information, establishing an interpretable cross-modal knowledge transfer paradigm.
\item We propose a bi-modal concurrent learning strategy to dynamically extract key features in WSIs and retrieved knowledge, effectively suppressing information redundancy while enhancing pathological semantic learning.
\item Experiments on the PathText (BRCA) dataset demonstrate that our method significantly outperforms existing benchmarks in report generation quality (BLEU-4=0.135, ROUGE-L=0.293) and Her-2 prediction metrics (F1=0.730).
\end{itemize}
\begin{figure}[t]
\includegraphics[width=\textwidth]{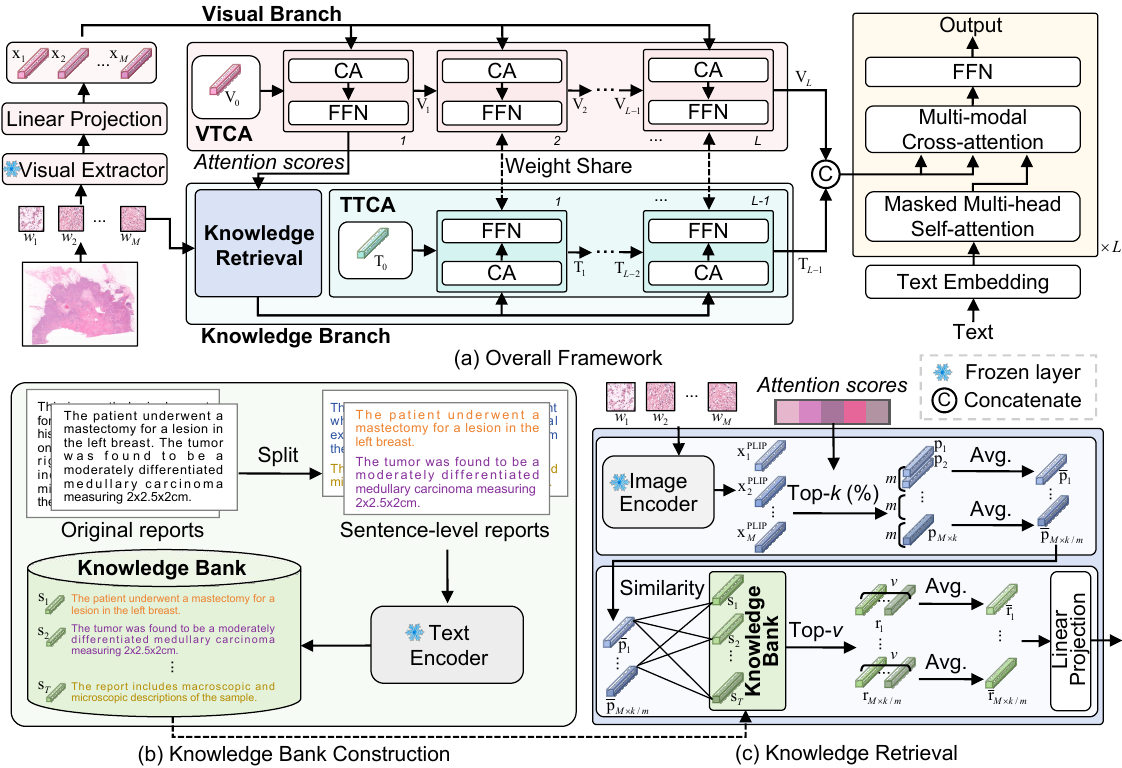}
\caption{Overview of BiGen. The overall framework in (a), consists of a bi-modal concurrent learning encoder and a multi-modal decoder. "CA" and "FFN" indicates Cross-attention and Feed-Forward Network. "VTCA" and "TTCA" refer to the visual and textual token cross-attention. We construct a knowledge bank in (b) and retrieve WSI-relevant knowledge from the knowledge bank through high-attention patches in (c).} 
\label{framework}
\end{figure}
\section{Method}
Our goal is to emulate pathologists’ diagnostic reasoning through the integration of visual evidence and historical
knowledge, to provide rich semantic content and suppress redundant information in WSIs. For this, we introduce a novel Historical Report Guided \textbf{Bi}-modal Concurrent Learning Framework for Pathology Report \textbf{Gen}eration (BiGen), which involves a visual branch and a knowledge branch (as illustrated in Fig.~\ref{framework} (a)). In the visual branch (Sec.~\ref{method:VTCA}) , compact visual features are iteratively aggregated through the visual token cross-attention (VTCA) module to focus on key pathological patterns. In the knowledge branch (Sec.~\ref{method:KR}), a knowledge retrieval module is designed to retrieve WSI-relevant knowledge features from a pre-constructed knowledge bank by matching high-attention patches from the visual branch. Furthermore, the retrieved knowledge features are aggregated through the textual token cross-attention (TTCA) module, sharing weights with the visual token cross-attention layers and obtaining rich semantic content. Finally, a multi-modal decoder (Sec.~\ref{method:decoder}) integrates both visual and textual tokens to generate comprehensive diagnostic reports.
\subsection{Problem Formulation}
The task of pathology report generation is formulated as a sequence-to-sequence problem, where the input consists of multiple image tokens extracted from a high-dimensional pathology image, and the output is a natural language report. Let \( \{ \mathbf{w}_i \}_{i=1}^{M} \) denote the set of non-overlapping patches cropped from the WSI \( \mathbf{W}\), where \( M\) is the number of patches. The objective is to generate the target pathology report \( \mathbf{Y} = \{ \mathbf{y}_n \}_{n=1}^{N} \), where \( N \) is the length of the report for $\mathbf{W}$.

A pretrained visual extractor is used to extract feature embeddings from \( \{ \mathbf{w}_i\}_{i=1}^{M} \), followed by a linear projection layer, resulting in a feature matrix \( \mathbf{X} = \{ \mathbf{x}_i \}_{i=1}^{M} \), where \( \mathbf{x}_i \in \mathbb{R}^{d} \) and \( d \) is the embedding dimension. The training objective is to maximize the likelihood of generating the correct report given the patch embeddings \( \mathbf{X} \). The total negative log-likelihood (NLL) loss is computed as: $
    \mathcal{L} = -  \sum_{n=1}^{N} \log P(\mathbf{y}_n | \{ \mathbf{y}_i \}_{i<n}, \mathbf{X})
$, where \( \{ \mathbf{y}_i \}_{i<n} \) represents the previously generated tokens in the report sequence.

\subsection{Visual Token Cross-attention}
\label{method:VTCA}
In the visual branch, we propose a visual token cross-attention module to extract a compact and informative visual representation of the WSI. Instead of performing self-attention over all patch features, we utilize a cross-attention mechanism, where a single learnable token attends to all patch features. Specifically, we initialize a visual token \( \mathbf{V}_0 \in \mathbb{R}^{1 \times d} \) as the query. The patch features \( \mathbf{X} \in \mathbb{R}^{M \times d} \) act as the key and the value, and the cross-attention mechanism is applied iteratively over \( L \) layers, yielding the following update rule:

\begin{equation}
    \mathbf{V}_{l} = \texttt{CrossAttn}(\mathbf{V}_{l-1}, \mathbf{X}; \mathbf{\Theta}_l), l \in \{1,2,\ldots,L\}, \label{visual_encoder}
\end{equation}

\noindent where \( \texttt{CrossAttn}(\cdot) \) denotes the cross-attention layer, and \( \mathbf{\Theta}_l \) means the $l$-th layer's learnable parameters. 

\subsection{Knowledge Branch}
Pathologists frequently recall historical diagnosis records to support diagnostic decisions. Inspired by this practice, we incorporate a knowledge retrieval mechanism to enhance the pathology report generation. PLIP \cite{huang2023visual} is a visual–language foundation model trained on over 200,000 paired pathology images and text data, allowing pathologists to search for similar cases through image or text queries, which is utilized to complete knowledge retrieval in this module.

\noindent\textbf{\underline{I. Knowledge Bank Construction:}}
Due to the text token length limitation, a long report can't be fed into the text encoder of PLIP \cite{huang2023visual}. So we construct a knowledge bank from sentence-level pathology reports to provide high-quality semantic knowledge, as illustrated in Fig.~\ref{framework} (b). Specifically, the original reports in the training set (\emph{e.g.}, PathText (BRCA)) are split into individual sentences and encoded using the text encoder of PLIP \cite{huang2023visual}. The resulting sentence embeddings are stored as the knowledge bank, denoted as \( \mathbf{S} = \{ \mathbf{s}_i \}_{i=1}^{T}\), where \( \mathbf{s}_i \in \mathbb{R}^{d} \), 
and \( T \) is the number of sentence embeddings in the knowledge bank.

\noindent\textbf{\underline{II. Knowledge Retrieval:}}
\label{method:KR}
As illustrated in Fig.~\ref{framework} (c), to retrieve knowledge using image embeddings aligned with sentence embeddings in the knowledge bank, we feed all patches \( \{ \mathbf{w}_i \}_{i=1}^{M} \) into the image encoder of PLIP \cite{huang2023visual} and obtain patch embeddings \( \mathbf{X}^\texttt{PLIP} = \{ \mathbf{x}^\texttt{PLIP}_i \}_{i=1}^{M} \), where \( \mathbf{x}^\texttt{PLIP}_i \in \mathbb{R}^{d} \) and \( d \) is the embedding dimension. Considering that many patches may be irrelevant to diagnosis, retrieving knowledge based on all patches could introduce redundant and noisy information. Inspired by \cite{chen2024image}, which optimizes computational efficiency by learning adaptive attention patterns in early layers and removing unimportant visual tokens in subsequent layers, we use the attention scores from the first layer of the visual token cross-attention module (Sec.~\ref{method:VTCA}) to identify key patches for knowledge retrieval. Specifically, the top-\( k \) patches with the highest attention scores are selected, where $k$ is the selecting ratio, and their corresponding image embeddings in \( \mathbf{X}^\texttt{PLIP} \) are collected as \( \mathbf{P} = \{ \mathbf{p}_i \}_{i=1}^{M\times k} \), where \( \mathbf{p}_i \in \mathbb{R}^{d} \).

Although top-$k$ patches are selected, similarity measure between selected patch embeddings and sentence embeddings still leads to massive calculation due to thousands of patch embeddings and thousands of sentence embeddings stored in the knowledge bank. Considering the spatial neighborhood patches with same tissue types on a WSI \cite{liu2021simtriplet}, we flatten the token sequence $\mathbf{P}$ and uniformly partition it into \( M\times k/m \) tissue regions with $m$ as the region size, ensuring that sparse patch embeddings are compressed. Then the average feature for each tissue region is computed as the region feature and updated tissue region embeddings are obtained: $\mathbf{\bar{P}} = \{ \mathbf{\bar{P}}_1, \mathbf{\bar{P}}_2, \dots, \mathbf{\bar{P}}_ {M\times k/m } \}$.

Next, the cosine similarity between each region's feature in $ \mathbf{\bar{P}}$ and all stored knowledge embeddings in the knowledge bank $\mathbf{S}$ are computed. For the $i$-th region, we select the top-\( v \) knowledge features based on the similarity and obtain $\mathbf{r}_i = \{ \mathbf{r}_i^j \}_{j=1}^{v}$, where \( \mathbf{r}_i^j \in \mathbb{R}^{d} \), and the average knowledge feature in $\mathbf{r}_i$ is computed as the retrieved knowledge feature $\mathbf{\bar{r}}_i$, where \( \mathbf{\bar{r}}_i \in \mathbb{R}^{d} \). All retrieved knowledge features are obtained: $\mathbf{R} = \{ \bar{\mathbf{r}}_i \}_{i=1}^{M\times k/m}$, followed by a linear projection layer.

\noindent\textbf{\underline{III. Textual Token Cross-attention:}}
To focus on key information from redundant retrieved knowledge, we introduce a learnable textual token \( \mathbf{T}_0 \in \mathbb{R}^{1 \times d} \). Similar to the visual token cross-attention module, we apply a cross-attention mechanism iteratively to refine the textual global token token:

\begin{equation}
    \mathbf{T}_{l} = \texttt{CrossAttn}(\mathbf{T}_{l-1}, \mathbf{R}; \mathbf{\Theta}_l), l \in \{1,2,\ldots,L-1\}.
    \label{retrival_encoder}
\end{equation}

To promote cross-modal alignment, the cross-attention layers of visual and knowledge branches share weights, obtaining rich semantic content. Since the knowledge retrieval relies on the attention scores from the first visual token cross-attention, the textual token cross-attention are deployed for $L-1$ layers.

\subsection{Multi-modal Decoder}
\label{method:decoder}
After refinement for both the visual and textual tokens, we concatenate the two tokens to form a joint representation: $\mathbf{F} = \texttt{{Concat}}(\mathbf{V}_L, \mathbf{T}_{L-1})$.

\begin{table}[t]
\centering
\renewcommand\arraystretch{0.5}
\caption{Results of pathology report generation on PathText (BRCA). BLEU-n: the BLEU score computed based on n-grams. \textbf{Bold}: the highest score. \underline{Underline}: 2nd score.}
\resizebox{0.9\linewidth}{!}{
\begin{tabular}{l |c c c c c c c| c c c}  
\toprule[1pt]
\multirow{2}{*}{Model} & \multicolumn{7}{c|}{NLP metrics}                           & \multicolumn{3}{c}{Classification metrics}                                                \\ \cline{2-11} 
                       & BLEU-1 & BLEU-2 & BLEU-3 & BLEU-4 & METEOR & ROUGE & Fact$_\text{ent}$ & \multicolumn{1}{c}{Precision} & Recall                         & F1\\ \hline
CNN-RNN \cite{vinyals2015show}&  0.371 &	0.185 &	0.089 &	0.043 	&0.143& 	0.239 & 0.478& 0.429& 0.750& 0.546 \\
att-LSTM \cite{pmlr-v37-xuc15}& 0.371&0.191 &0.094 &0.048 &0.142 &0.238 &0.460&\underline{0.600}&0.636&\underline{0.618} \\
vanilla Transformer \cite{vaswani2017attention}& 0.389  & 0.246  & 0.157  & 0.103  & 0.158  & 0.257  & 0.487 &0.457 &	\underline{0.800} &	0.582
 \\
R2Gen \cite{chen2020generating}& 0.378  & 0.243  & 0.160  & 0.107  & \underline{0.179}  & \underline{0.279}  & 0.505 & 0.486 &	0.630 &	0.548 
\\
R2GenCMN \cite{chen2022cross}& 0.396  & 0.254  & 0.164  & 0.110  & 0.163  & \underline{0.279}  & 0.470 & 0.429 &	0.682 &	0.526 
\\
MI-Gen \cite{chen2024wsicaption}& 0.416  & 0.267  & 0.174  & 0.115  & 0.165  & 0.270  & 0.490 & 0.257 &	0.529 &	0.346 
\\
HistGen \cite{guo2024histgen}& \underline{0.422}  & \underline{0.272}  & \underline{0.177}  & \underline{0.118}  & 0.169  & 0.277  & \underline{0.493} & 0.514 &	0.643 &	0.571 
\\
ours & \textbf{0.450}  & \textbf{0.296 } & \textbf{0.196}  & \textbf{0.135}  & \textbf{0.180}  & \textbf{0.293}  & \textbf{0.521} & \textbf{0.657} &	\textbf{0.821} &	\textbf{0.730} 
\\
\bottomrule[1pt] 
\end{tabular}}
\label{tab_result}
\end{table}

The report sequence is generated auto-regressively, implemented by Masked Multi-head Self-attention. Specifically, in the $l$-th decoder layer, the previously generated words $\{ \mathbf{y}_i^l \}_{i<n}$ serve as queries, the fused representation \( \mathbf{F} \) provides keys and values for the decoder and the $n$-th word $\mathbf{y}_{n}^l$ is predicted as:

\begin{equation}
    \mathbf{y}_{n}^l = \texttt{CrossAttn}(\{ \mathbf{y}_i^l \}_{i<n}, \mathbf{F}; \mathbf{\Omega}_l),l \in \{1,2,\ldots,L\},
\end{equation}

\noindent where \( \mathbf{\Omega}_l \) indicates learnable weights for \( l \)-th cross-attention layer in the decoder.

\begin{figure}[t]
\centering
\includegraphics[width=0.9\textwidth]{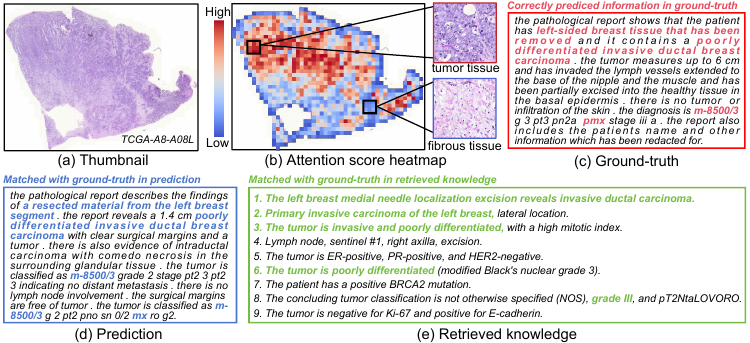}
\caption{Visualization of the (a) thumbnail, (b) attention score heatmap, (c) report ground-truth, (d) prediction and (e) retrieved knowledge of sample "TCGA-A8-A08L".
} 
\label{wsi_report}
\end{figure}
\section{Experiments and Results}
\subsection{Implementation Details} 
\noindent\textbf{\underline{I. Datasets:}}
Following \cite{chen2024wsicaption}, we conduct experiments on PathText-BRCA with the same data splits for training, validation, and testing sets. Additionally, we remove duplicate patient samples across these sets, ensuring no patient overlap. Then the training, validation, and test sets contain 796, 88, and 93 samples, respectively.

\noindent\textbf{\underline{II. Model Setting:}}
Following \cite{chen2024wsicaption}, WSIs are processed via CLAM \cite{lu2021data} to extract non-overlapping 256×256 tissue patches at 10x magnification, the number of encoder and decoder layers are both 3 with 4 attention heads and the embedding size is 512. UNI \cite{chen2024towards} (pretrained on over 100 million tissue patches and over 100,000 WSIs) serves as the visual extractor. Besides, in the knowledge branch, we set $k=0.4$, $m=20$, $v=3$. We employ Adam optimizer with initial learning rate 1e-4 and weight decay 5e-5. We adopt beam search with the size of 3 as the sampling method. All experiments is conducted on a single A40-40G GPU. 
\begin{table}[t]
\centering
\renewcommand\arraystretch{0.3}
\caption{Ablation study on key components. WS denotes weight sharing between the visual and knowledge branch. WSL denotes weight sharing across layers in each branch. VTCA/TTCA refer to the visual/textual token cross-attention. W/o VTCA or TTCA refers to conducting self-attention. KR refers to the knowledge retrieval module. "AVG. $\triangle$" represents the average metric promotion compared to the vanilla Transformer.}
\resizebox{0.9\linewidth}{!}{
\begin{tabular}{ccccc|cccccccc}
\toprule[1pt]
WS & WSL & VTCA & KR& TTCA & BLEU-1 & BLEU-2 & BLEU-3 & BLEU-4 & METEOR & ROUGE & Fact$_\text{ent}$  & AVG. $\triangle$\\ \hline
\ding{55}  & \ding{55} & \ding{55}   & \ding{55}      & \ding{55}         & 0.389  & 0.246  & 0.157  & 0.103  & 0.158  & 0.257 & 0.487 & \textendash \\
\ding{55}  & \ding{55} & \ding{51}   & \ding{55}      & \ding{55}         & 0.436  & 0.283  & 0.186  & 0.126  & 0.174  & 0.287 &    0.499  &\textbf{10.84\%} \\
\ding{55}  & \ding{51} & \ding{51}   & \ding{55}      & \ding{55}         & 0.437  & 0.286  & 0.190  & 0.127  & 0.174  & 0.280 &    0.507  &\textbf{11.41\%} \\
\ding{55}  & \ding{51}   & \ding{51}      & \ding{51}  & \ding{55}        & 0.416  & 0.269  & 0.176  & 0.119  & 0.169  & 0.283 &    0.505  &\textbf{7.75\%} \\
\ding{55}  & \ding{51}   & \ding{51}      & \ding{51}  & \ding{51}        & 0.441  & 0.288  & 0.190  & 0.130  & 0.178  & 0.290 &    0.515  &\textbf{13.06\%} \\
\ding{51}  & \ding{51}   & \ding{51}      & \ding{51}  & \ding{51}        & \textbf{0.450}  & \textbf{0.296}  & \textbf{0.196}  & \textbf{0.135}  & \textbf{0.180}  & \textbf{0.293} & \textbf{0.521} &\textbf{15.26\%} \\
\bottomrule[1pt] 
\end{tabular}}
\label{ablation}
\end{table}
\subsection{Comparison between Ours and Other Methods}
We compare with seven image captioning methods in Table~\ref{tab_result}: two LSTM-based methods: CNN-RNN \cite{vinyals2015show} and att-LSTM \cite{pmlr-v37-xuc15}, five Transformer-based methods: Transformer \cite{vaswani2017attention}, R2Gen \cite{chen2020generating} and R2GenCMN \cite{chen2022cross} (both designed for radiology image captioning), MI-Gen \cite{chen2024wsicaption} and HistGen \cite{guo2024histgen} (both designed for pathology report genaration). Following \cite{chen2024wsicaption}, we adopt four Natural Language Processing (NLP) metrics to evaluate generation performance: BLEU \cite{papineni2002bleu}, METEOR \cite{banerjee2005meteor}, ROUGE \cite{rouge2004package} and Fact$_\text{ent}$ \cite{miura2020improving}. Besides, we adopt three classification metrics to evaluate Her-2 prediction in generated reports: Precision, Recall and F1 score.

Table~\ref{tab_result} shows LSTM-based methods underperform Transformer variants in NLP metrics. R2Gen \cite{chen2020generating} and R2GenCMN \cite{chen2022cross} fall short in recognizing key visual features when applied to WSI analysis. MI-Gen \cite{chen2024wsicaption} feed visual tokens into Transformer to model relationships between patches, which dilutes critical pathological features due to the overwhelming number of redundant regions. Although the local-global hierarchical encoding strategy of HistGen \cite{guo2024histgen} improves efficiency, it still struggles to select key pathological features and lack semantic content. Our historical report guided bi-modal concurrent learning method provides rich semantic content and distills critical pathological features and knowledge features iteratively, achieving 0.135 in BLEU-4 (+14.4\% versus HistGen \cite{guo2024histgen}). 

Fig.~\ref{wsi_report} (b) visualizes the attention scores from the 1st layer in the visual token cross-attention module, highlighting tumor tissues versus fibroblast tissues, indicating key feature extraction capability of our method. Fig.~\ref{wsi_report} (c-e) demonstrate our method effectively generates medical terms consistent with the ground truth, \emph{e.g.}, ``m-8500/3'', a morphological term in ICD-O-3 \cite{fritz2000international}, indicating invasive ductal carcinoma. Additionally, retrieved knowledge enhances semantics, \emph{e.g.}, ``Primary invasive carcinoma of the left breast'', improving diagnostic accuracy.

\begin{figure}[t]
\centering
\includegraphics[width=0.9\textwidth]{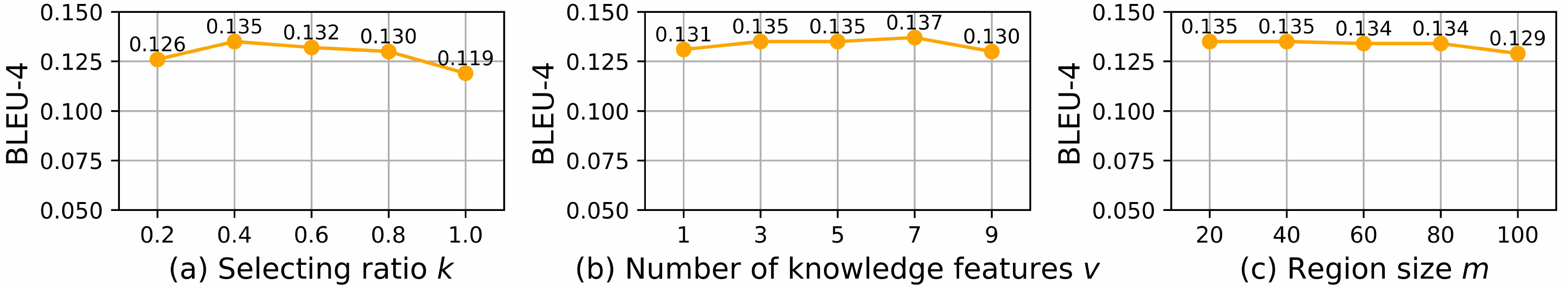}
\caption{Performance changes by varying the (a) selecting ratio $k$, (b) number of knowledge features $v$ and region size $m$.}
\label{parameter}
\end{figure}
\subsection{Ablation Study}
We conduct ablation studies on five components: 1)Weight Sharing (WS) between the visual and knowledge branch, 2) Weight Sharing across Layers (WSL) in each branch, 3) Visual Token Cross-attention (VTCA), 4) Knowledge Retrieval (KR) and 5) Textual Token Cross-attention (TTCA). As shown in Table~\ref{ablation}, the baseline model without any proposed modules (\emph{i.e.}, vanilla Transformer) achieves the lowest performance (Row 1 in Table~\ref{ablation}), confirming the necessity of addressing redundancy of WSIs and lack of semantic content. Introducing VTCA (Row 2 in Table~\ref{ablation}) and TTCA (Row 5 in Table~\ref{ablation}) yield over 10\% and 5\% relative improvements, representatively. This demonstrates our bi-modal concurrent learning strategy effectively mitigates information redundancy in both WSIs and retrieved knowledge. Enabling WSL (Row 3 in Table~\ref{ablation}) further enhances BLEU-4, which suggests weight sharing in each branch stabilizes feature propagation. Simply incorporating the KR module (Row 4 in Table~\ref{ablation}) causes performance drop with redundant knowledge. However, by employing TTCA (Row 5 in Table~\ref{ablation}), consistent performance gain across all metrics, attributed to recognizing key knowledge embeddings with valuable semantic content. Notably, sharing weights between two branches (Row 6 in Table~\ref{ablation}) achieves optimal performance via cross-modal visual-knowledge alignment. 

Fig.~\ref{parameter} shows three hyper-parameter sensitivity, with default values $k=0.4$, $v=3$ and $m=20$. As illustrated in Fig.~\ref{parameter} (a), it can be seen that $k$ is not sensitive in $[0.4,0.8]$. When $k$ is too large, \emph{i.e.}, $k=1$, which means using all patch embeddings to retrieve knowledge, irrelative patches brings considerable redundant information, resulting in obvious performance degradation. As illustrated in Fig.~\ref{parameter} (b), the best performance can be achieved when $v$ changes from 3 to 7. As illustrated in Fig.~\ref{parameter} (c), performance keeps stable when $m$ varies with the range of $[20,80]$, which maintains the best performance with lower computation.
\section{Conclusion}
In this paper, we present a novel historical report guided bi-modal concurrent learning framework for pathology report generation to address two challenges: lack of semantic content and considerable redundancy in WSIs. The proposed knowledge retrieval provides semantically rich diagnostic report information and the bi-modal concurrent learning strategy extracts key pathological features and key knowledge features. Extensive experiments on the PathText (BRCA) dataset demonstrate state-of-the-art performance both in report generation and Her-2 prediction. This work establishes a new paradigm for pathology report generation which effectively integrates visual evidence and historical knowledge and mitigates information redundancy in both WSIs and retrieved knowledge, with potential applications extending to other medical image interpretation tasks.

\begin{credits}
\subsubsection{Acknowledgements} This work was supported by the National Natural Science Foundation of China (Grant No. 62471182), {Shanghai Rising-Star Program (Grant No. 24QA2702100)}, and the Science and Technology Commission of Shanghai Municipality (Grant No. 22DZ2229004).

\subsubsection{Disclosure of Interests.} The authors have no competing interests to declare that are relevant to the content of this article.
\end{credits}
\bibliographystyle{splncs04}
\bibliography{Paper-2364.bib}
\end{document}